\documentclass[runningheads]{llncs}
\usepackage[T1]{fontenc}
\usepackage{tikz, pgfplots, algorithmic, algorithm, url, amsmath}
\usepackage{graphicx}

\usepackage{hyperref}

\makeatletter
\newcommand{\printfnsymbol}[1]{%
  \textsuperscript{\@fnsymbol{#1}}%
}
\makeatother

\begin{document}

\title{The Swiss Army Knife for Image-to-Image Translation: \\ Multi-Task Diffusion Models}

\titlerunning{The Swiss Army Knife for Image-to-Image Translation}

\author{Julia Wolleb\thanks{equal contribution}, Robin Sandk\"uhler\printfnsymbol{1}, Florentin Bieder, Philippe C. Cattin}

\authorrunning{J. Wolleb et al.}

\institute{Department of Biomedical Engineering, University of Basel, Allschwil, Switzerland\\
\email{julia.wolleb@unibas.ch}}
\maketitle              
\begin{abstract}
Recently, diffusion models were applied to a wide range of image analysis tasks. We build on a method for image-to-image translation using denoising diffusion implicit models and include a regression problem and a segmentation problem for guiding the image generation to the desired output. The main advantage of our approach is that the guidance during the denoising process is done by an external gradient. Consequently, the diffusion model does not need to be retrained for the different tasks on the same dataset. We apply our method to simulate the aging process on facial photos using a regression task, as well as on a brain magnetic resonance (MR) imaging  dataset for the simulation of brain tumor growth. Furthermore, we use a segmentation model to inpaint tumors at the desired location in healthy slices of brain MR images. We achieve convincing results for all problems.
\keywords{Diffusion models \and Image-to-image translation \and Regression \and Segmentation.}
\end{abstract}

\section{Introduction}
\begin{figure}[h!]
    \centering
    \includegraphics[width=0.95\textwidth]{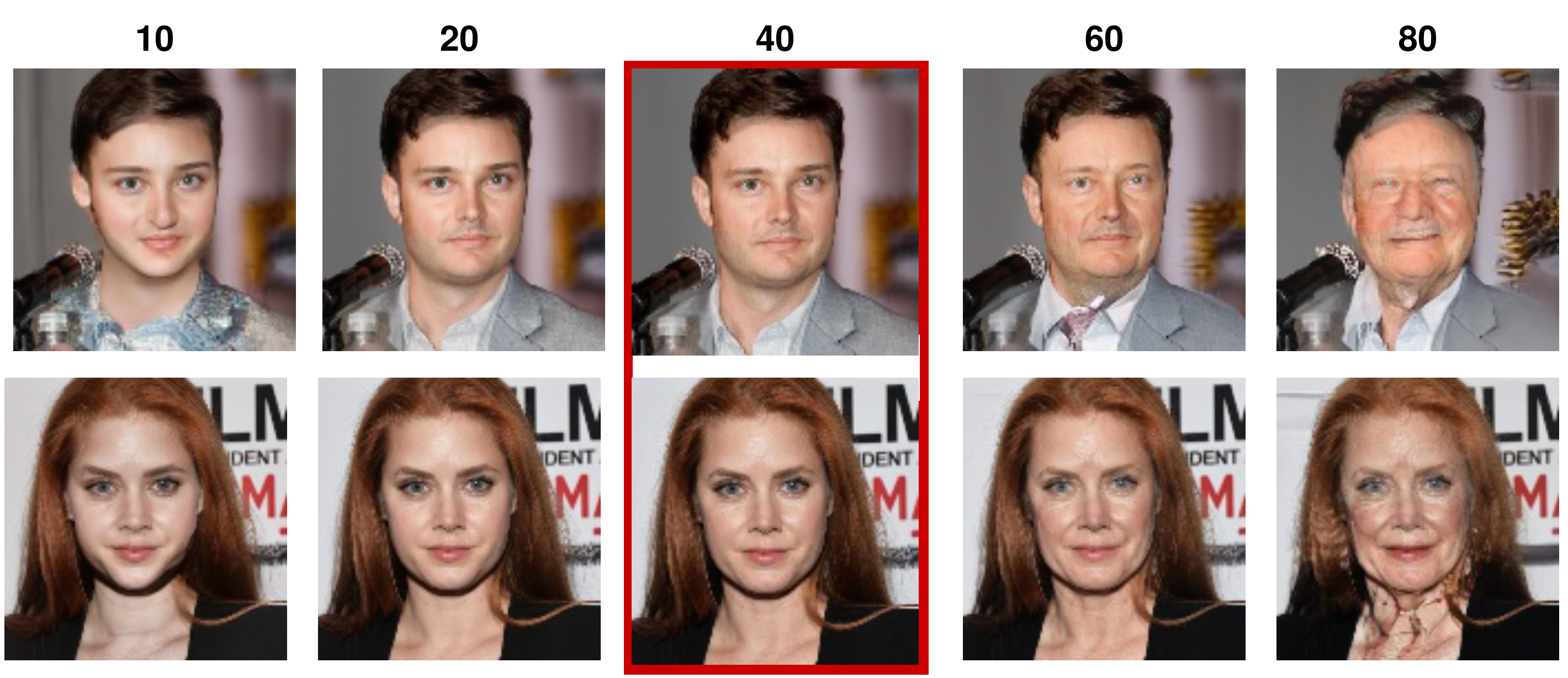}
    \caption{Results of our method for the age regression task on facial photos. The original images framed in red show exemplary subjects of age 40. With our approach, they are translated to images matching a desired age value $i \in \{10,20,60,80\}$.  }
    \label{dev}
\end{figure}

For many applications, it is of great interest to perform image-to-image translation such that the output image is changed in some regions to the desired characteristics  and unchanged in other regions, e.g., the image background.
\begin{figure}[h!]
    \centering
    \includegraphics[width=0.9\textwidth]{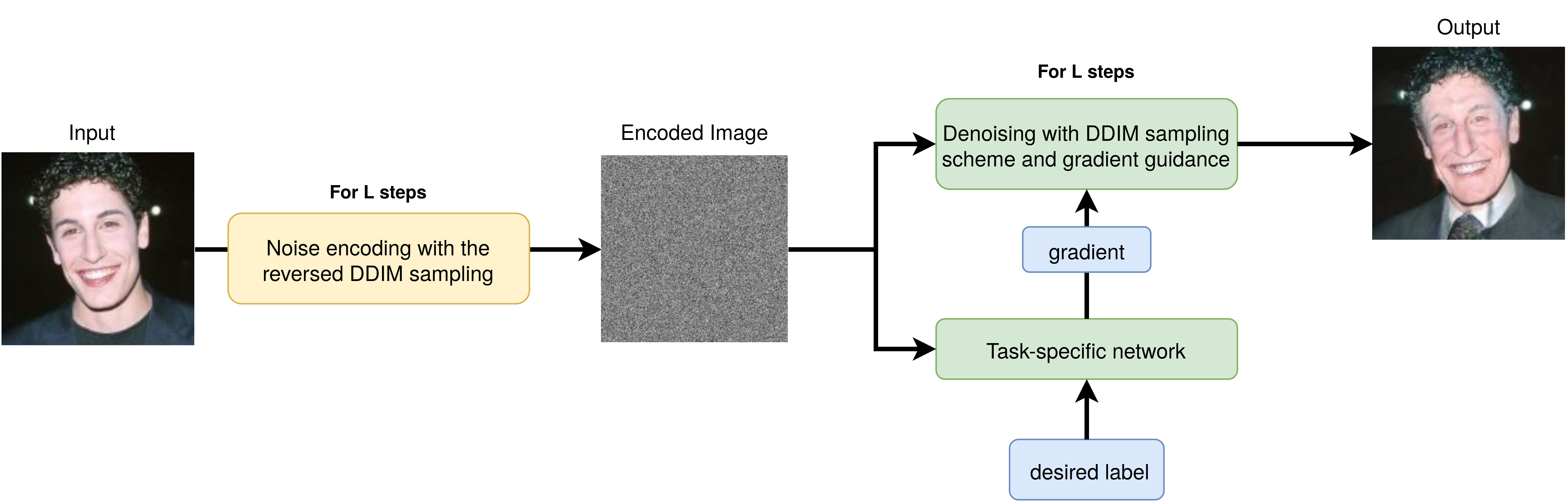}
    \caption{Proposed scheme for image-to-image translation for the example of age regression.}
    \label{overview}
\end{figure}
 Most approaches rely on generative adversarial nets (GANs) \cite{gan} or variational autoencoders \cite{pang2021image}.
However, the adversarial training of GANs can be difficult and requires a lot of hyperparameter tuning. 
 Furthermore, a big challenge is that only image features related to the desired output characteristics should be changed.
To circumvent those issues, we propose an approach based on
 Denoising Diffusion Probabilistic Models (DDPMs)\cite{ddpm} and the sampling scheme of Denoising Diffusion Implicit Models (DDIMs) \cite{ddim}. \\
 In \cite{wolleb2022anomaly}, we performed image-to-image translation between different classes by training a DDPM, and using the DDIM noising and denoising scheme and classifier guidance during sampling.
We build on this approach and present a method for image-to-image translation for variable tasks. 
Figure~\ref{overview} shows an overview of the proposed approach. First, we separately train a DDPM as well as an external task-specific model on the same dataset. In this work, this external model is a regression or a segmentation model. The case of a classification model is already described in \cite{wolleb2022anomaly}. Image-to-image translation is performed only during the sampling process.
For this, we first encode the information about the input image with the iterative noising process of DDIMs. 
During the denoising process using the DDIM sampling scheme, we inject the gradient of the task-specific model in each sampling step.
By scaling this gradient, the generation of the output image is guided towards the desired output. \\
For the regression problem, we apply our algorithm to a dataset of facial images \cite{kaggle} for age prediction, where image generation is guided towards a desired age. An example for this is given in Figure~\ref{dev}, where we translate photos of two exemplary people of age $40$ to photos showing the same person at various other ages.  Moreover, we use a regression task on the BRATS2020 challenge \cite{brats1,brats2,brats3} for the simulation of brain tumor growth. The segmentation problem is applied to the BRATS2020 dataset for the translation of images of healthy subjects to images containing a brain tumor at a location that we can freely choose.

\subsubsection{Related Work}
In computer vision, image-to-image translation towards specific image attributes has been a task of great interest and includes style transfer \cite{liu2021multiple}, relighting or colorization tasks \cite{santhanam2017generalized}, and changing facial attributes such as hair color or gender \cite{choi2018stargan}.
The translation of images to subjects of another age has been explored by using autoencoders \cite{lan2021age} based on facial landmarks, or by adapting GANs \cite{huang2020pfa,sharma2021prediction}.  A big challenge of those approaches is that only task-related features should be changed, and the rest of the image remains consistent with the input image.\\
Lately, DDPMs were in focus due to their success in tasks such as image generation \cite{beatgans}, image-to-image translation \cite{unitddpm,ilvr}, segmentation \cite{diffseg,implicitens}, reconstruction \cite{palette} and registration\cite{diffusemorph}. In \cite{rasul2021autoregressive}, denoising diffusion models were used for multivariate probabilistic time series forecasting. In \cite{liu2021more}, the gradient of image-text or image matching scores are used to guide the generation of synthetic images.\\
The presented method is based on our previous work  \cite{wolleb2022anomaly}, where image-to-image translation was performed based on a binary classification problem. This approach used classifier guidance during the sampling process, such that the training of the DDPM is not changed and a pretrained model can be used. The big advantages are the straightforward training process, and the fact that only features related to the classification problem are changed. The rest of the image is preserved.\\
In \cite{preechakul2021diffusion}, diffusion autoencoders were proposed for meaningful representation learning. They encode image information in a latent space and use a conditional DDIM to manipulate the image attributes. Very recently, \cite{su2022dual} proposed the training of two diffusion models for translations between arbitrary pairs of source-target domains. In contrast to our method, those approaches depend on a manipulation of the latent space, and the diffusion models need to be retrained  for each application. For our method, the same diffusion model can be used for various applications, as the guidance towards the desired attributes is done only during the sampling process by an external gradient.

\section{Method}\label{method}
The goal is to perform image-to-image translation such that the output image matches task-specific criteria. Thereby, it is important that only features related to the task are changed, and all other features are preserved. 
Our method follows \cite{wolleb2022anomaly}, where the task-specific criteria is given by a binary classifier. In this work, we adapt this method with a regression model as described in Algorithm \ref{alg:reg}, as well as with a segmentation model, as described in Algorithm \ref{alg:seg}. \\
We implement a DDPM according to  \cite{ddpm,improving}. 
For an input image $x$,  we add small amounts of noise for many steps $T$, such that we get a series of increasingly noisy images $\{x_0, x_1, ..., x_T\}$. A diffusion model is given by a U-Net $\epsilon_{\theta}$, which is trained with the MSE loss to predict  $x_{t-1}$ from $x_t$. During sampling, a synthetic image $x_0$ can be generated from  $x_T \sim \mathcal{N}(0,\mathbf{I})$ by predicting $x_{t-1}$ from $x_{t}$ for $t \in \{T,...,1\}$ using \eqref{ddim2}.
During training, we can explicitly write the forward noising process as
\begin{equation}\label{eq:property}
x_{t}=\sqrt[]{\overline{\alpha} _{t}}x_{0}+\sqrt[]{1-\overline{\alpha} _{t}}\epsilon, \quad \mbox{with } \epsilon \sim \mathcal{N}(0,\mathbf{I}).
\end{equation}
Here, we define $\alpha _{t}:=1-\beta _{t}$ and $\overline{\alpha}_{t}:=\prod_{s=1}^t \alpha _{s}$, where  $\beta_{1},...,\beta_{T}$ denote the forward process variances. 
This noisy image $x_t$ given in \eqref{eq:property} serves as input for the U-Net $\epsilon_{\theta}$, which is trained using the MSE loss 
\begin{equation}\label{eq4}
\mathcal{L}:= ||\epsilon-\epsilon_{\theta}(\sqrt[]{\overline{\alpha} _{t}}x_{0}+\sqrt[]{1-\overline{\alpha} _{t}}\epsilon, t)||^2_2 , \quad \mbox{with } \epsilon \sim \mathcal{N}(0,\mathbf{I}).
\end{equation}
We can then to predict $x_{t-1}$ from $x_t$ with
\begin{equation}\label{ddim2}
x_{t-1} = \sqrt{\alpha_{t-1}}\left(\frac{x_t-\sqrt{1-\alpha_{t}}\epsilon_\theta(x_t,t)}{\sqrt{\alpha_t}}\right)+\sqrt{1-\alpha_{t-1}-\sigma_t^2}\epsilon_{\theta}(x_t,t)+\sigma_t\epsilon. 
\end{equation}
 In DDIMs, we set ${\sigma_t=0}$, which results in a deterministic sampling process.
 Equation \eqref{ddim2} can be interpreted as the 
Euler method to solve the ordinary differential equation (ODE) described in \cite{ddim}. By solving the reversed ODE, we can reverse the generation process. Consequently, using enough discretization steps,  we can encode $x_{t+1}$ given $x_t$ with
\begin{equation}\label{reversed}
x_{t+1}  =  x_{t}+\sqrt{\bar{\alpha}_{t+1}} \left[ \left( \sqrt{\frac{1}{\bar{\alpha}_{t}}} - \sqrt{\frac{1}{\bar{\alpha}_{t+1}}}\right) x_t + \left(\sqrt{\frac{1}{\bar{\alpha}_{t+1}} - 1} - \sqrt{\frac{1}{\bar{\alpha}_{t} }- 1} \right) \epsilon_\theta(x_t,t) \right].
\end{equation}
For image-to-image translation, we define a noise level $L \in \{1, ... , T\}$, as proposed in \cite{wolleb2022anomaly}.
By applying \eqref{reversed} for $t \in \{0,...,L-1\}$, we can encode an image $x_0$ in a noisy image $x_L$. With this deterministic, iterative noising process, the information of the input image $x$ is stored in $x_L$. \\
In addition to the diffusion model, we train a separate task-specific network  on our set of noisy images $\{x_0, x_1, ..., x_T\}$.  During the denoising process, we follow \eqref{ddim2} with $\sigma_t=0$ for $t \in \{L, ... , 1\}$, and add the gradient of the task-specific network in every step during sampling  to lead the image generation to the desired output characteristics. The case of classification was already presented in \cite{wolleb2022anomaly}. In Sections~\ref{methodR} and \ref{methodS}, we investigate the tasks of regression and segmentation.

\subsection{Regression}\label{methodR}
The regression model $R$ follows the architecture of the encoder of the diffusion model, and is trained with the MSE loss.
 The iterative noising and denoising scheme for image-to-image translation for a regression problem is presented in Algorithm \ref{alg:reg}. We encode an image $x$ in a noisy image $x_L$, and define a desired value $i$. 
 During the denoising process, the gradient $\nabla_{x_t}  R(x_t,t)$ of the regression model is used to update $\epsilon_{\theta}(x_t,t)$. The sign of the gradient defines the direction in which the generation process is influenced, e.g., whether the subject is made younger or older.\\
 To guide the image generation to the desired value $i$ for the regression task, we define the gradient scale \mbox{$s_t=i- R(x_t,t)$}, such that the influence of the gradient gets smaller if the predicted value is close to the desired value. If the predicted value surpasses $i$, the sign of $s_t$ is changed, and the denoising process is guided back in the other direction. An additional gradient scale $c$ is constant over time and can be used to further amplify the gradient, as we explored in \cite{wolleb2022anomaly}.

\begin{algorithm}
    \caption{Regression guidance}
    \label{alg:reg}
    \begin{algorithmic}
        \STATE Input: input image $x$, desired value $i$, noise level $L$, constant gradient scale $c$\\
        Output: synthetic image $x_{0}$
            \STATE $x_0 \gets x$
            \FORALL{$t$ from 0 to $L-1$ }
            \STATE $x_{t+1} \gets x_{t}+\sqrt{\bar{\alpha}_{t+1}} \left[ \left( \sqrt{\frac{1}{\bar{\alpha}_{t}}} - \sqrt{\frac{1}{\bar{\alpha}_{t+1}}}\right) x_t + \left(\sqrt{\frac{1}{\bar{\alpha}_{t+1}} - 1} - \sqrt{\frac{1}{\bar{\alpha}_{t} }- 1} \right) \epsilon_\theta(x_t,t) \right]$
            \ENDFOR
        \FORALL{$t$ from $L$ to 1}
           \STATE $s_t\gets i- R(x_t,t)$
            \STATE $\hat \epsilon \gets \epsilon_{\theta}(x_t,t) - s_t\*c \*\sqrt{1-\bar{\alpha}_t}  \nabla_{x_t}  R(x_t,t)$
            \STATE $x_{t-1} \gets \sqrt{\bar{\alpha}_{t-1}} \left( \frac{x_t - \sqrt{1-\bar{\alpha}_t} \hat{\epsilon}}{\sqrt{\bar{\alpha}_t}} \right) + \sqrt{1-\bar{\alpha}_{t-1}} \hat{\epsilon}$
        \ENDFOR
   
        \RETURN $x_{0}$
    \end{algorithmic}
\end{algorithm}
\newpage
\subsection{Segmentation}\label{methodS}
The algorithm for segmentation guidance can be found in Algorithm~\ref{alg:seg}.
 The iterative noise encoding scheme stays the same as in Section~\ref{methodR}. However, instead of a regression network, we train a segmentation network $S$ on the noisy images $\{x_0, x_1, ..., x_T\}$ with the cross-entropy loss. The architecture of the segmentation model follows the U-Net architecture of the diffusion model. We further define a desired label mask $z$, and compute the binary cross-entropy loss $H$ between the output of the segmentation network $S(x_t,t)$ and a desired label mask $z$. We can then guide the image generation towards an image that matches $z$ by using the gradient  $\nabla_{x_t} H$ during the denoising process. As already proposed in Section~\ref{methodR}, a constant gradient scale $c$ can be applied to amplify this gradient.

\begin{algorithm}
    \caption{Segmentation guidance}
    \label{alg:seg}
    \begin{algorithmic}
        \STATE Input: input image $x$, desired label mask $z$, noise level $L$, constant gradient scale $c$, number of pixels $P$\\
        Output: synthetic image $x_{0}$
            \STATE $x_0 \gets x$
            \FORALL{$t$ from 0 to $L-1$ }
            \STATE $x_{t+1} \gets x_{t}+\sqrt{\bar{\alpha}_{t+1}} \left[ \left( \sqrt{\frac{1}{\bar{\alpha}_{t}}} - \sqrt{\frac{1}{\bar{\alpha}_{t+1}}}\right) x_t + \left(\sqrt{\frac{1}{\bar{\alpha}_{t+1}} - 1} - \sqrt{\frac{1}{\bar{\alpha}_{t} }- 1} \right) \epsilon_\theta(x_t,t) \right]$
            \ENDFOR
        \FORALL{$t$ from $L$ to 1}
           \STATE $H \gets   -\frac{1}{P} \sum\limits_{j=1}^{P} \bigl(z_j\log S(x_t,t)_j+(1-z_j)\log (1-S(x_t,t)_j)\bigr)$
            \STATE $\hat \epsilon \gets \epsilon_{\theta}(x_t,t) - c \* \sqrt{1-\bar{\alpha}_t}  \nabla_{x_t}  H$
            \STATE $x_{t-1} \gets \sqrt{\bar{\alpha}_{t-1}} \left( \frac{x_t - \sqrt{1-\bar{\alpha}_t} \hat{\epsilon}}{\sqrt{\bar{\alpha}_t}} \right) + \sqrt{1-\bar{\alpha}_{t-1}} \hat{\epsilon}$
        \ENDFOR
   
        \RETURN $x_{0}$
    \end{algorithmic}
\end{algorithm}

\section{Experiments}
The DDPM is trained as proposed in \cite{improving} without any data augmentation.
We choose $T=1000$, $L=400$. The other hyperparameters for the DDPM are described in the appendix of \cite{beatgans}. The model is trained with the Adam optimizer and the hybrid loss objective described in \cite{improving}, with a learning rate of ${10^{-4}}$, and a batch size of 10. The number of channels in the first layer is chosen as 128, and using one attention head at resolution 16. The regression model has a depth of $4$ and uses attention heads at the resolution of $8, 16$ and $32$. We set the number of channels in the first layer to 32 for the BRATS2020 dataset, and to 128 for the dataset of facial photos.
Training was performed on an NVIDIA Quadro RTX 6000 GPU, with Pytorch 1.7.1 as software framework.

\subsubsection{Facial Photos}
The dataset of facial photos shows people aged between 1 and 100. The training dataset comprises 185,631 images of size $3 \times 128\times 128$. The test set includes 47,568 images. All images are normalized to values between $0$ and $1$.
The diffusion model is trained for  500,000 iterations, due to the high variability in the data. 
The regression model is trained for 80,000 iterations. The number of parameters is 85,606,150 for the diffusion model, and  67,061,121 for the regression model. Image-to-image translation for one image takes $73$ s for the age regression task.

\subsubsection{BRATS2020}\label{brats}
The BRATS2020 dataset of 3D brain Magnetic Resonance (MR) images of subjects with a brain tumor is described in \cite{wolleb2022anomaly}. We only consider 2D axial slices. For each slice, four different MR sequences as well as the pixel-wise ground truth segmentation of the tumor is given. The images are of size $4 \times 256\times 256$, where each channel shows one of the four MR sequences, namely T1-weighted, T2-weighted, FLAIR, and T1-weighted with contrast enhancement (T1ce). All images  are normalized to values between 0 and 1.  For the regression problem, the relative tumor size is calculated as the ratio between the tumor size and the brain size for each slice. Our training set includes 16,205 slices, whereas there are 1,787 images in the test set. 
We train the diffusion model for 50,000 iterations, the regression and segmentation models for 20,000 iterations. The number of model parameters is 113,681,160 for the diffusion and the segmentation model, and 5,452,833 for the regression model. Image-to-image translation for one image takes $84$ s for the regression task and $102$ s for the segmentation task.

\section{Results and Discussion}

\subsection{Age Regression on Facial Photos}\label{face}
\begin{figure}[h!]
    \centering
    \includegraphics[width=\textwidth]{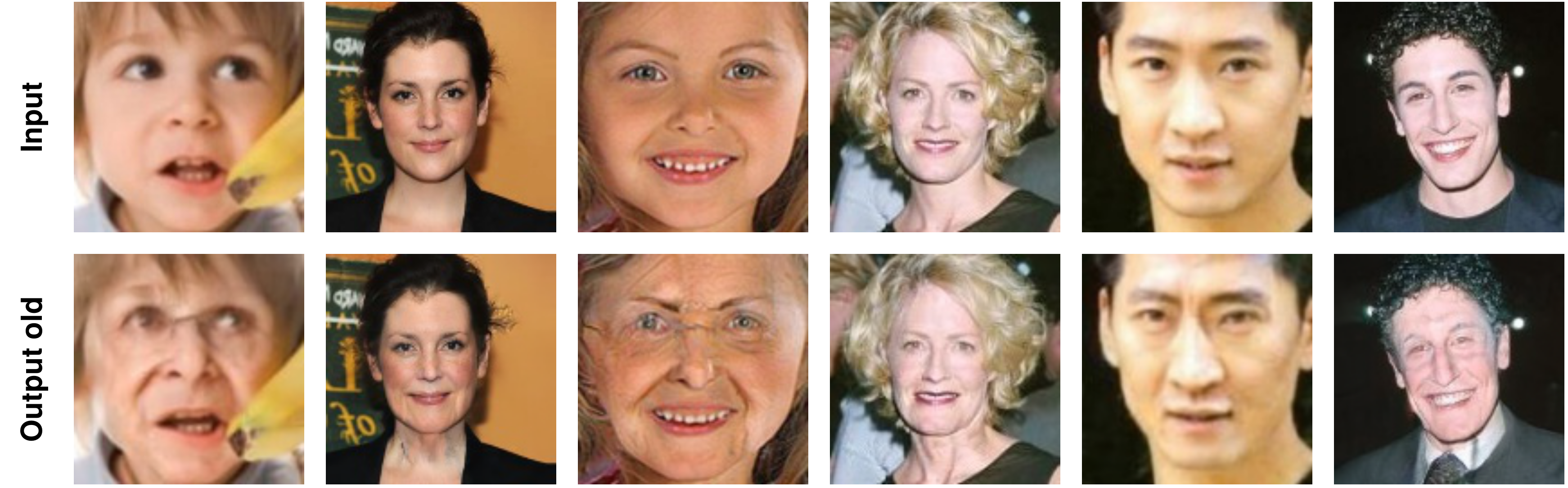}
    \caption{Results of our method for $s_t=1$. The positive gradient makes the subjects older.}
    \label{makeold}
\end{figure}

\begin{figure}[h]
    \centering
    \includegraphics[width=\textwidth]{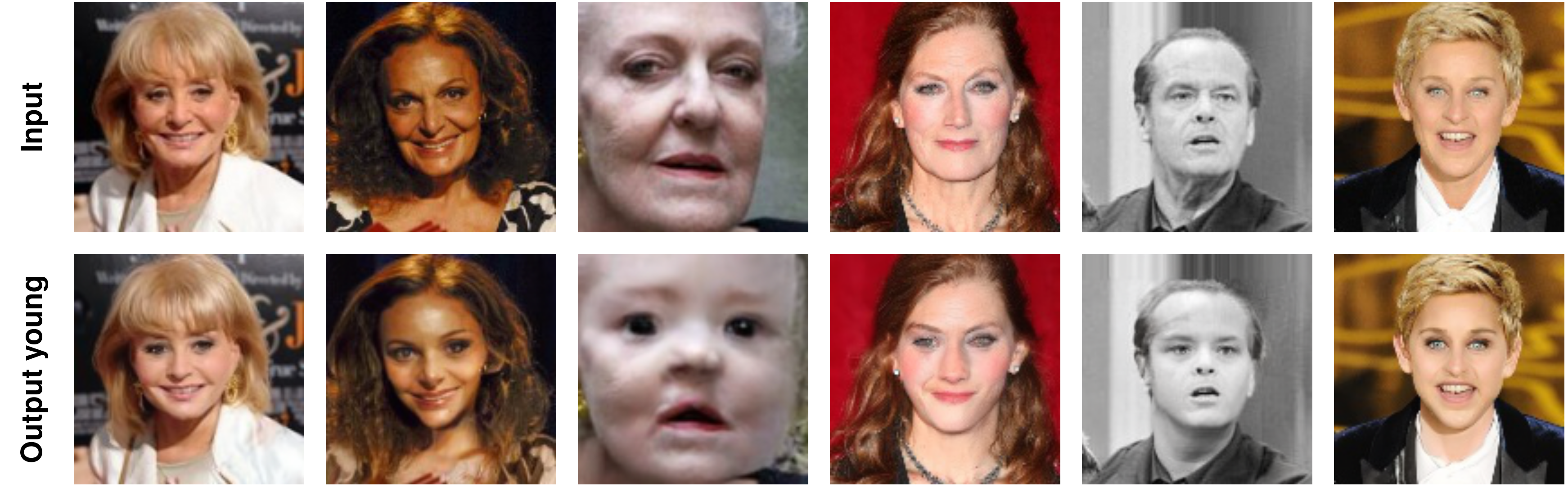}
    \caption{Results of our method for $s_t=-1$. The negative gradient makes the subjects younger. }
    \label{makeyoung}
\end{figure}

In Figure~\ref{makeold}, we present exemplary images of the test set, as well as the output of our model, if we set $s_t=1 \quad \forall{t}$. This results in aging of the subjects. In Figure~\ref{makeyoung}, we present further examples, as well as the output of our model, if we set $s_t=-1 \quad \forall{t}$. We see that the output images show younger subjects, whereas the background and other features such as hair and clothes are preserved.  For this dataset, we choose $c=5$.\\
As described in Section~\ref{methodR}, we set  $s_t=i- R(x_t,t)$ if we wish to generate an image of the desired age $i$. In Figure~\ref{dev}, we show the aging process for two subjects of age 40, and set the desired age values to  $i \in \{10,20,60,80\}$. 

\newpage
\subsection{Regression on the Relative Tumor Size} \label{regressiontumor}
\begin{figure}[h!]
    \centering
    \includegraphics[width=\textwidth]{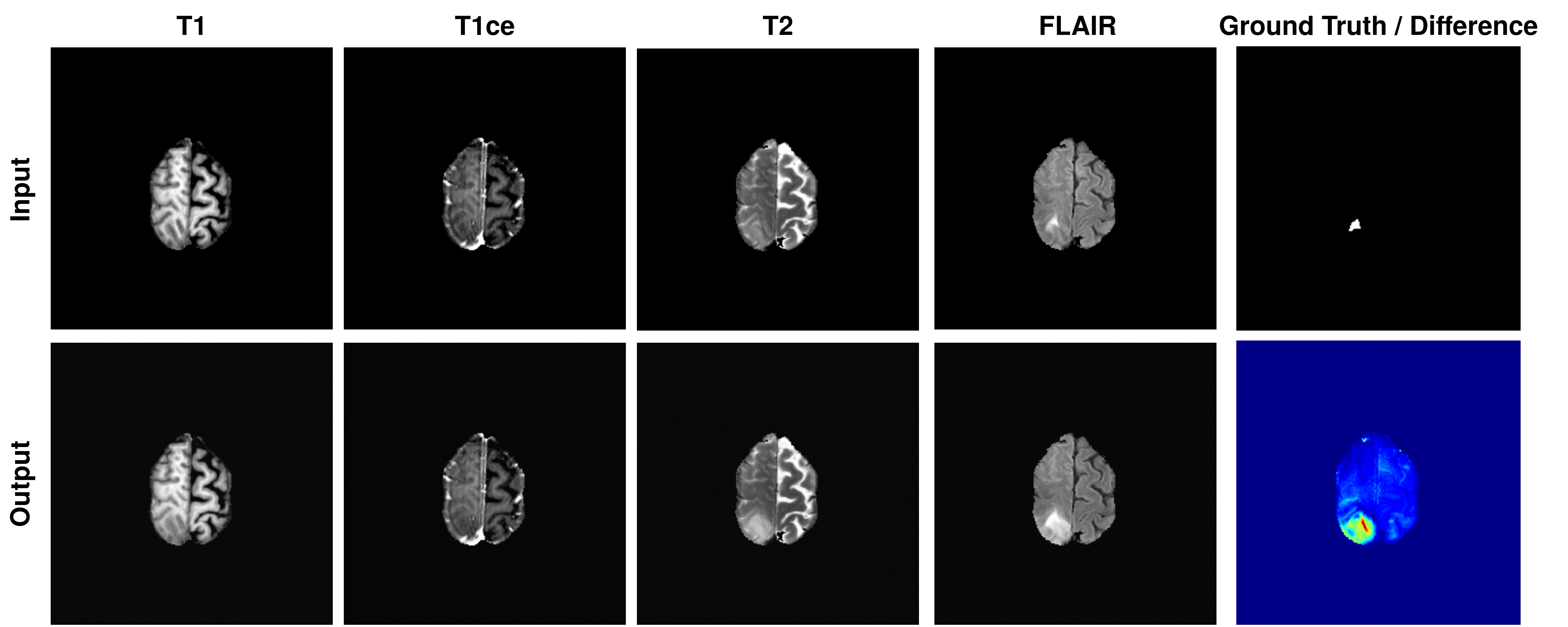}
    \caption{Results of our method for $s_t=1$, which results in an enlarged tumor. }
    \label{grow}
\end{figure}

\begin{figure}[h!]
    \centering
    \includegraphics[width=\textwidth]{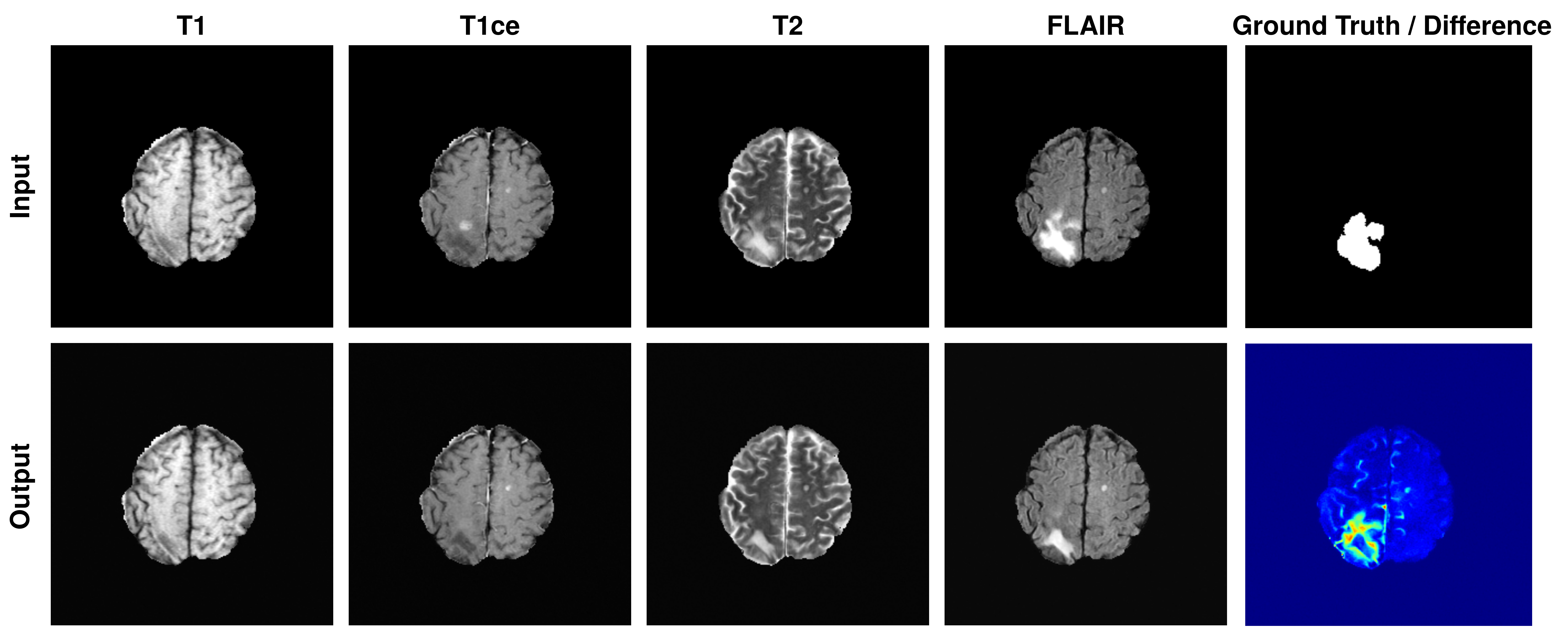}
    \caption{Results of our method for $s_t=-1$, which leads to a smaller tumor. }
    \label{decline}
\end{figure}

\begin{figure}[h!]
    \begin{center}
    \includegraphics[width=\textwidth]{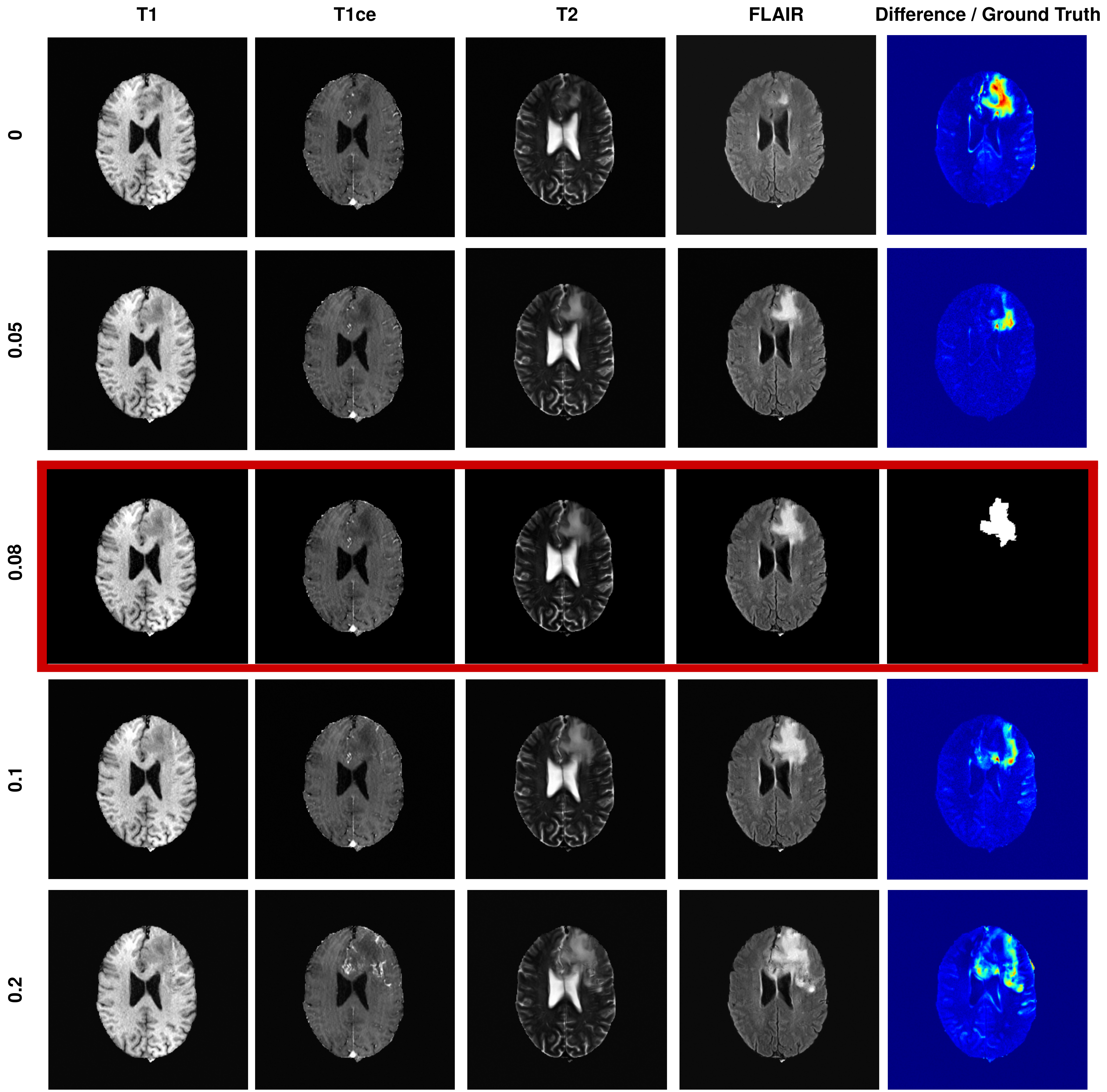}
    \caption{Results of our method for $s_t=i- R(x_t,t)$ and $i \in \{0, 0.05,0.1, 0.2\}$. The original image framed in red has a ratio of $0.08$. The difference maps in the last column highlight the regions that are changed during image-to-image translation.}
    \label{progression}
    \end{center}
\end{figure}

We train a regression model on the BRATS2020 dataset described in Section~\ref{brats}. The value for each slice is defined as the ratio of the area of the ground truth segmentation mask and the area of the brain. Like in Section~\ref{face}, we can make the tumor grow or shrink by changing the sign of the gradient. In Figure~\ref{grow}, we present the input and output images for all four MR sequences for $s_t=1$. The output shows images with an enlarged tumor. This can also be seen on the difference map in the last column, where the absolute difference between the input and the output image, summed over all $4$ channels, is presented. On the other hand, Figure~\ref{decline} shows the output of our method for $s_t=-1$, resulting in a smaller tumor. On the BRATS2020 dataset, we choose $c=1000$.\\
Just like in the age regression problem, we can influence the tumor size by providing a desired value $i$. In Figure~\ref{progression}, we show the results for various desired values \mbox{$i \in \{0, 0.05,0.1, 0.2\}$}, where the original value is $0.08$. All four MR sequences as well as the absolute difference map between the input and the output image are provided.

\newpage
\subsection{Tumor Generation using Segmentation Models}\label{segmentation}
\begin{figure}[h]
    \centering
    \includegraphics[width=\textwidth]{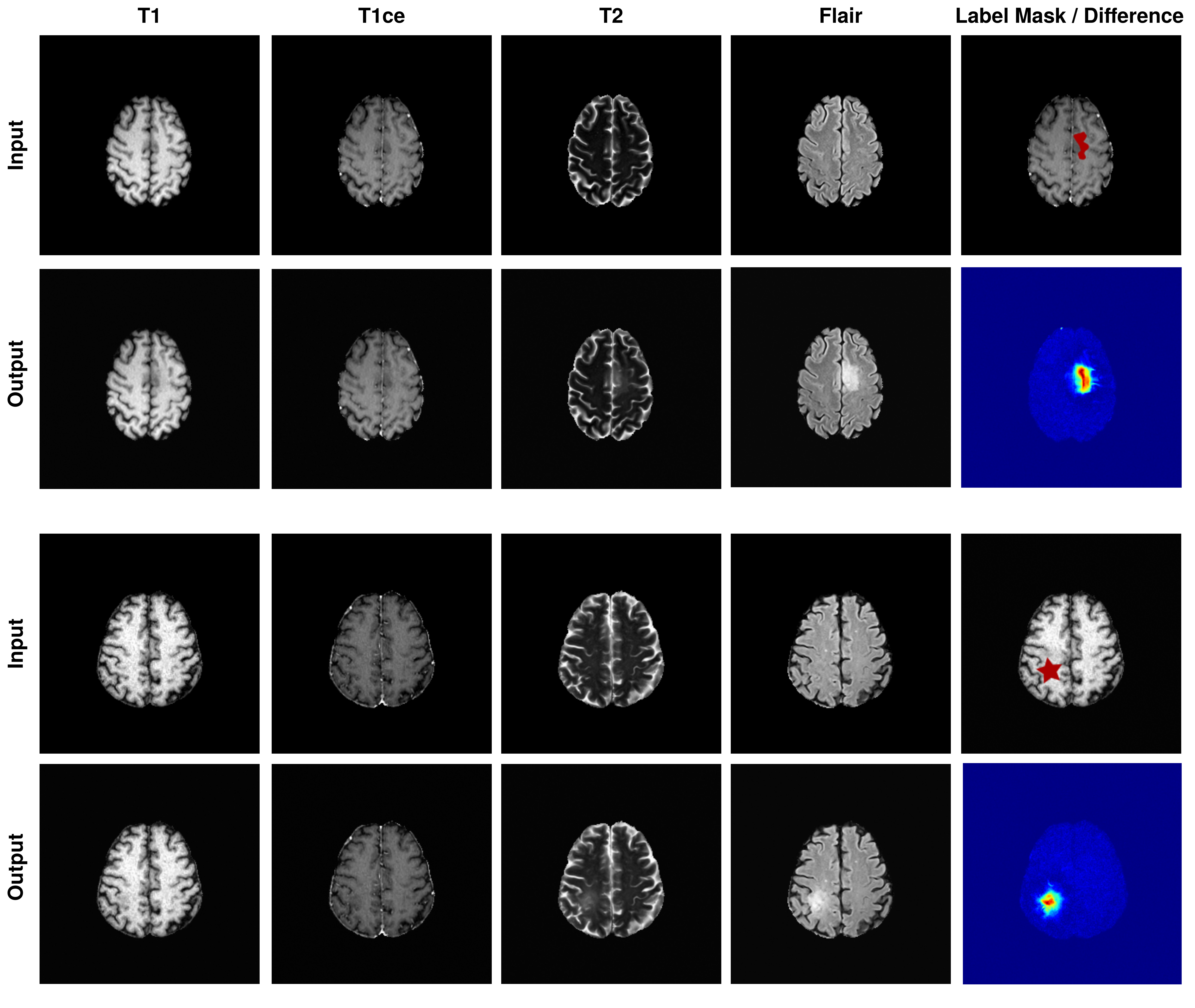}
    \caption{Results for the segmentation task on the BRATS2020 dataset. The label mask is shown in the last column in red, together with one channel of the input image as a reference in the background. The difference map shows that a tumor was inpainted in the desired area. }
    \label{segtumor}
\end{figure}

We train a fully supervised segmentation model on the BRATS2020 dataset for brain tumor segmentation. For image-to-image translation, we aim to translate an image showing a healthy slice into a slice containing a tumor. For this, we define a pixel-wise label mask where we want the model to insert a tumor. \\
In the first row of Figure~\ref{segtumor}, we present the input image showing a healthy slice, as well as the desired label mask in red. The output image shows a brain MR image containing a fake tumor.
By considering the difference between the input and output image, we see that this fake tumor was drawn in the desired area. For these experiments, we set $c=5$. This approach could be helpful for the generation of artificial data  for the training of anomaly detection methods.

\section{Conclusion}
In this paper, we present an image-to-image translation method based on DDIMs for different specific tasks using gradient guidance. In addition to a classical DDPM, a separate network for the specific task is trained. We show that our method can translate images to an output matching a desired value or label mask of a regression or segmentation problem. The resulting images are only altered in features related to that desired characteristics, and the rest of the image is preserved. The big advantage of this approach is that the same diffusion model can be used for multiple tasks, i.e., classification, segmentation or regression. \\
We applied our method to a dataset of facial photos for age regression, and on the BRATS2020 dataset for brain tumor growth. By using the segmentation task, we are able to insert brain tumors in healthy slices of the BRATS2020 dataset. This can be useful for the generation of artificial training data for anomaly detection methods. We achieve convincing results on both datasets for both tasks. Future work includes speeding up the sampling process, and the extension to 3D data.

\subsubsection{Acknowledgements} This research was supported by the Novartis FreeNovation initiative and the Uniscientia
Foundation (project $\#$147-2018).

%
\newpage
\bibliographystyle{splncs04}
\bibliography{adddim}

\begin{thebibliography}{10}
\providecommand{\url}[1]{\texttt{#1}}
\providecommand{\urlprefix}{URL }
\providecommand{\doi}[1]{https://doi.org/#1}

\bibitem{brats2}
Bakas, S., Akbari, H., Sotiras, A., Bilello, M., Rozycki, M., Kirby, J.S.,
  Freymann, J.B., Farahani, K., Davatzikos, C.: Advancing the cancer genome
  atlas glioma {MRI} collections with expert segmentation labels and radiomic
  features. Scientific data  \textbf{4}(1),  1--13 (2017)

\bibitem{brats3}
Bakas, S., Reyes, M., Jakab, A., Bauer, S., Rempfler, M., Crimi, A., Shinohara,
  R.T., Berger, C., Ha, S.M., Rozycki, M., et~al.: Identifying the best machine
  learning algorithms for brain tumor segmentation, progression assessment, and
  overall survival prediction in the {BRATS} challenge. arXiv preprint
  arXiv:1811.02629  (2018)

\bibitem{diffseg}
Baranchuk, D., Rubachev, I., Voynov, A., Khrulkov, V., Babenko, A.:
  Label-efficient semantic segmentation with diffusion models. arXiv preprint
  arXiv:2112.03126  (2021)

\bibitem{ilvr}
Choi, J., Kim, S., Jeong, Y., Gwon, Y., Yoon, S.: Ilvr: Conditioning method for
  denoising diffusion probabilistic models. arXiv preprint arXiv:2108.02938
  (2021)

\bibitem{choi2018stargan}
Choi, Y., Choi, M., Kim, M., Ha, J.W., Kim, S., Choo, J.: Stargan: Unified
  generative adversarial networks for multi-domain image-to-image translation.
  In: Proceedings of the IEEE conference on computer vision and pattern
  recognition. pp. 8789--8797 (2018)

\bibitem{beatgans}
Dhariwal, P., Nichol, A.: Diffusion models beat gans on image synthesis.
  Advances in Neural Information Processing Systems  \textbf{34} (2021)

\bibitem{kaggle}
Frențescu, M.: Age prediction,
  \url{https://www.kaggle.com/datasets/mariafrenti/age-prediction}, retrieved
  26.02.2022

\bibitem{gan}
Goodfellow, I., Pouget-Abadie, J., Mirza, M., Xu, B., Warde-Farley, D., Ozair,
  S., Courville, A., Bengio, Y.: Generative adversarial nets. Advances in
  neural information processing systems  \textbf{27} (2014)

\bibitem{ddpm}
Ho, J., Jain, A., Abbeel, P.: Denoising diffusion probabilistic models.
  Advances in Neural Information Processing Systems  \textbf{33},  6840--6851
  (2020)

\bibitem{huang2020pfa}
Huang, Z., Chen, S., Zhang, J., Shan, H.: Pfa-gan: Progressive face aging with
  generative adversarial network. IEEE Transactions on Information Forensics
  and Security  \textbf{16},  2031--2045 (2020)

\bibitem{diffusemorph}
Kim, B., Han, I., Ye, J.C.: Diffusemorph: Unsupervised deformable image
  registration along continuous trajectory using diffusion models. arXiv
  preprint arXiv:2112.05149  (2021)

\bibitem{lan2021age}
Lan, L.C., Liu, T.J., Liu, K.H.: Age regression with specific facial landmarks
  by dual discriminator adversarial autoencoder. In: 2021 IEEE International
  Conference on Image Processing (ICIP). pp. 2718--2722. IEEE (2021)

\bibitem{liu2021more}
Liu, X., Park, D.H., Azadi, S., Zhang, G., Chopikyan, A., Hu, Y., Shi, H.,
  Rohrbach, A., Darrell, T.: More control for free! image synthesis with
  semantic diffusion guidance. arXiv preprint arXiv:2112.05744  (2021)

\bibitem{liu2021multiple}
Liu, Z.S., Kalogeiton, V., Cani, M.P.: Multiple style transfer via variational
  autoencoder. In: 2021 IEEE International Conference on Image Processing
  (ICIP). pp. 2413--2417. IEEE (2021)

\bibitem{brats1}
Menze, B.H., Jakab, A., Bauer, S., Kalpathy-Cramer, J., Farahani, K., Kirby,
  J., Burren, Y., Porz, N., Slotboom, J., Wiest, R., et~al.: The multimodal
  brain tumor image segmentation benchmark ({BRATS}). IEEE transactions on
  medical imaging  \textbf{34}(10),  1993--2024 (2014)

\bibitem{improving}
Nichol, A.Q., Dhariwal, P.: Improved denoising diffusion probabilistic models.
  In: Proceedings of the 38th International Conference on Machine Learning.
  vol.~139, pp. 8162--8171. PMLR (2021)

\bibitem{pang2021image}
Pang, Y., Lin, J., Qin, T., Chen, Z.: Image-to-image translation: Methods and
  applications. IEEE Transactions on Multimedia  (2021)

\bibitem{preechakul2021diffusion}
Preechakul, K., Chatthee, N., Wizadwongsa, S., Suwajanakorn, S.: Diffusion
  autoencoders: Toward a meaningful and decodable representation. arXiv
  preprint arXiv:2111.15640  (2021)

\bibitem{rasul2021autoregressive}
Rasul, K., Seward, C., Schuster, I., Vollgraf, R.: Autoregressive denoising
  diffusion models for multivariate probabilistic time series forecasting. In:
  International Conference on Machine Learning. pp. 8857--8868. PMLR (2021)

\bibitem{palette}
Saharia, C., Chan, W., Chang, H., Lee, C.A., Ho, J., Salimans, T., Fleet, D.J.,
  Norouzi, M.: Palette: Image-to-image diffusion models. arXiv preprint
  arXiv:2111.05826  (2021)

\bibitem{santhanam2017generalized}
Santhanam, V., Morariu, V.I., Davis, L.S.: Generalized deep image to image
  regression. In: Proceedings of the IEEE Conference on Computer Vision and
  Pattern Recognition. pp. 5609--5619 (2017)

\bibitem{unitddpm}
Sasaki, H., Willcocks, C.G., Breckon, T.P.: Unit-ddpm: Unpaired image
  translation with denoising diffusion probabilistic models. arXiv preprint
  arXiv:2104.05358  (2021)

\bibitem{sharma2021prediction}
Sharma, N., Sharma, R., Jindal, N.: Prediction of face age progression with
  generative adversarial networks. Multimedia Tools and Applications
  \textbf{80}(25),  33911--33935 (2021)

\bibitem{ddim}
Song, J., Meng, C., Ermon, S.: Denoising diffusion implicit models. arXiv
  preprint arXiv:2010.02502  (2020)

\bibitem{su2022dual}
Su, X., Song, J., Meng, C., Ermon, S.: Dual diffusion implicit bridges for
  image-to-image translation. arXiv preprint arXiv:2203.08382  (2022)

\bibitem{wolleb2022anomaly}
Wolleb, J., Bieder, F., Sandk{\"u}hler, R., Cattin, P.C.: Diffusion models for
  medical anomaly detection. arXiv preprint arXiv:2203.04306  (2022)

\bibitem{implicitens}
Wolleb, J., Sandk{\"u}hler, R., Bieder, F., Valmaggia, P., Cattin, P.C.:
  Diffusion models for implicit image segmentation ensembles. arXiv preprint
  arXiv:2112.03145  (2021)

\end{thebibliography}

\end{document}